\relax
\documentclass[letterpaper]{article} 
\usepackage{aaai21}  
\usepackage{times}  
\usepackage{helvet} 
\usepackage{courier}  
\usepackage[hyphens]{url}  
\usepackage{graphicx} 
\usepackage{natbib}  
\usepackage{caption} 
\usepackage{amsmath}
\usepackage{amssymb}
\usepackage{amsfonts}  
\usepackage{tikz}
\usepackage{caption}
\usepackage{subcaption}
\usepackage{color}
\usepackage{bm}
\title{Perturbation-based exploration methods in deep reinforcement learning }
\author{
    Sneha Aenugu\\
}
\affiliations{
    University of Massachusetts Amherst\\

   saenugu@cs.umass.edu

}

\begin{document}
\newcommand{\greenline}{\raisebox{2pt}{\tikz{\draw[-,black!40!green,solid,line width = 1.5pt](0,0) -- (8mm,0);}}}
\newcommand{\blueline}{\raisebox{2pt}{\tikz{\draw[-,black!0!blue,solid,line width = 1.5pt](0,0) -- (8mm,0);}}}
\newcommand{\orangeline}{\raisebox{2pt}{\tikz{\draw[-,orange,solid,line width = 1.5pt](0,0) -- (8mm,0);}}}

\maketitle

\begin{abstract}
Recent research on structured exploration placed emphasis on identifying novel states in the state space and incentivizing the agent to revisit them through intrinsic reward bonuses. In this study, we question whether the performance boost demonstrated through these methods is indeed due to the discovery of structure in exploratory schedule of the agent or is the benefit largely attributed to the perturbations in the policy and reward space manifested in pursuit of structured exploration. In this study we investigate the effect of perturbations in policy and reward spaces on the exploratory behavior of the agent. We proceed to show that simple acts of perturbing the policy just before the softmax layer and introduction of sporadic reward bonuses into the domain can greatly enhance exploration in several domains of the arcade learning environment. In light of these findings, we recommend benchmarking any enhancements to structured exploration research against the backdrop of noisy exploration.
\end{abstract}

\noindent 
\vspace{1mm}

\section{Introduction}
The exploration-exploitation trade-off is a concept at the heart of reinforcement learning. As an agent cannot afford to explore all the actions at every crossroad of the Markovian decision step, the question of how to explore efficiently to maximize its return from the domain comes into prominence. Therein lies the appeal of structured exploration, to systematically identify which actions are worth exploring and which states are worth revisiting.

Most of the research on structured exploration centered around the notion of curiosity-driven exploration where relatively novel states are given a disproportionate preference for exploration. This exploratory bias is injected into the environment by awarding intrinsic reward bonuses to those states that are deemed novel and promising. Several studies quantified the novelty of a state through estimation of pseudo-counts of state-visitation (\cite{count-based}, \cite{unifying}, \cite{successor}, \cite{hash}), training prediction models of state-visitation dynamics (\cite{dpn}, \cite{icm}, \cite{vime}, \cite{disagree}, \cite{rnd}). In fact \cite{curiosity} showed that intrinsic rewards generated through novelty detection are enough to drive the exploration in complex domains in the absence of any extrinsic rewards from the environment.

While the research in structured exploration has demonstrated enhanced exploration in several Atari 2600 games including several hard-exploration domains like Montezuma's revenge, the question of how much of this enhancement can be legitimately attributed to the \emph{structured} exploration comes into the forefront. In fact \cite{bonus} claimed that improved performance in some of the domains maybe better attributed to architectural changes rather than better to exploration schemes. In this study, we contend that the perturbations caused in the reward space through the injection of intrinsic reward bonuses contribute partly to the improved exploration. To validate this claim, instead of injecting reward bonuses through density/prediction models, we generate random reward bonuses and introduce them sporadically into the environment. We proceed to show that even these sporadic reward bonuses can greatly enhance the exploration in several domains of arcade learning environment.

We also extend the analysis of perturbations in the reward space to the policy space in policy gradient algorithms. Inspired by the structured reward shaping through the introduction of intrinsic rewards, we describe a method to shape policies to encourage selective exploratory behavior. We then contrast the structured policy shaping with the perturbed version where noise is injected into the policy network just before the softmax layer. We show that perturbations in policy space provide remarkable improvements in exploration in domains with sparse rewards.

We restrict our analysis to the policy gradient algorithms although some of the techniques can be easily translated to other algorithms like DQN. For baselines, we chose to work with Proximal Policy Optimization (PPO), Actor Critic using Kronecker-Factored Trust Region (ACKTR), Advantage Actor Critic (A2C) algorithms, which we then augment with perturbation-based exploration techniques to provide evidence for improvement in agent's performance. 

\section{Exploration in RL}
In this section, we review the existing literature on exploration in reinforcement learning (RL). We categorize the techniques of exploration into two broad categories:
\begin{enumerate}
    \item \emph{Structured exploration}: There is a formulation of criteria for the agent's selective preference towards a particular action or state.  
    \item \emph{Randomized exploration}: There is no structure in the criteria for exploration and any preference for a particular action/state is a result of randomization.
\end{enumerate}
\subsection{Structured exploration}
In these class of exploration techniques, the agent's past history in the environment and the structure of the state space is considered to formulate a plan for the agent to efficiently explore the domain.  

In this technique, the extrinsic reward from the environment is augmented by intrinsic rewards to selectively attend to the states which are either hitherto unexplored or are likely to yield high future rewards. The total reward function is thus represented as $r_t = r^{ext} + \beta r^{int}$ where $\beta$ is the factor modulating the trade-off between exploration and exploitation.

  \subsubsection{Estimation of state-visitation counts through density models} 
    \cite{unifying} constructed a density model to calculate the pseudo-counts for each state to estimate the agent's state-visitation frequency. Let $\rho(s)$ be the density model of the state s after visiting a sequence of states $s_{1:n}$. Let $\rho'(s)$ be the density model after the next occurrence of s. Given the pseudo-count of the state s as $N(s)$, then 
    \begin{equation}
        \rho(s) = \frac{N(s)}{n}; \rho'(s) = \frac{N(s)+1}{n}
    \end{equation}
    where $n$ is the normalization constant. Then the pseudo-count $N(s)$ can be expressed in terms of $\rho(s)$ and $\rho'(s)$ as
    \begin{equation}
        N(s) = \frac{\rho(s)(1-\rho'(s))}{\rho'(s)-\rho(s)}
    \end{equation}
    and the intrinsic reward is given by
    \begin{equation}
        r^{int}(s) = (N(s))^{-1/2}.
    \end{equation}
    \cite{unifying} used a Context Tree Switching (\cite{cts}) density model to capture pseudo-counts. \cite{count-based}, the other hand, used Pixel-CNN (\cite{pixelcnn}) as the density model.
    \subsubsection{Estimation of a prediction model}
    \cite{ims} introduced the notion of intrinsic motivation system which uses a forward-dynamics prediction model to identify the states which are novel to the agent and motivate the agent to visit those states with greater preference. 

    In this class of algorithms, the agent learns a dynamic model of the environment by learning to predict the next state from the current state and action, $f:(\phi(s_t), a_t) \rightarrow \phi(s_{t+1})$, where $\phi$ is the state space encoding function. The prediction error $e(\phi(s_t), a_t) = \Vert f(\phi(s_t),a_t) - \phi(s_{t+1}) \Vert_2^2$ can be used to drive the exploration, as greater the error the more novel the state is and therefore needs to be prioritized. The intrinsic reward proportional to the prediction error $r^{int} = \frac{e(s_t,a_t)}{t.C}$ is provided to the agent, where $C$ is the decay constant. The encoding function $\phi$ is learned in different ways (\cite{wengexp}):
    \begin{enumerate}
        \item Identity function: No encoding function is used, raw image pixels are used as input.
        \item Autoencoder is used for encoding state features (\cite{dpn}).
        \item A random network is used to encode the state space (\cite{rnd}).
        \item A self-supervised inverse dynamics model (\cite{icm}).
    \end{enumerate}
    
\subsubsection{Exploration through entropy maximization}
Entropy regularization is a popular technique to increase the exploratory performance of the agent. It prevents the agent from being stuck in a local minima by enabling it to explore all the available actions. This technique essentially serves to ensure that the predictability of the agent's choices is low,  thus resulting in increased entropy. By employing an entropy term in the loss function (\cite{deepenergy}), the policy updates tend to gravitate towards those policies where the randomness of the actions are preserved. Although the process of entropy regularization is aimed at improving randomness, we characterize it as a technique in structured exploration as it strives to ensure that no action in a given state is given a stronger preference over the rest of the actions.

\subsection{Randomized exploration}
In this class of techniques, any exploration schedule followed by the agent is a result of randomization. Noise is introduced into the action space or the policy space to encourage the randomness in the agent's choice resulting in an improvement in exploration.
\subsubsection{Action-space noise}
The trivial case of perturbing the action space for exploration is $\epsilon$-greedy or Boltzmann sampling of policy in the context of Q learning (\cite{qlearning}). As the Q-learning algorithm chooses the action with the highest Q-value, exploration is induced in the algorithm by introducing perturbations in the action space. Perturbing the policy parameters for exploration in policy gradient algorithms is first introduced in \cite{sde}, where a state-dependent offset is added to an action at each time step which maintains the variability of exploratory randomness in a state across episodes but will ensure that the offset is same for every visit of the state within an episode.  A state-dependent function $\hat{\epsilon}(\mathbf{s}, \hat{\mathbf{\theta}})$ is added to the policy to generate the action, $a = \pi(\mathbf{s}) + \hat{\epsilon}(\mathbf{s}, \hat{\mathbf{\theta}}) $ where the randomness is provided by the sampled parameters $\hat{\mathbf{\theta}} \sim \mathcal{N}(0, \sigma^2)$. 
\subsubsection{Parameter space noise}
\cite{noisy}, \cite{pnoise} introduced noise into the network parameters of the agent's policy. It was demonstrated that this added stochasticity aided exploratory performance of the agent by enhancing the performance in several domains of Arcade learning environment in both on-policy algorithms like TRPO (\cite{trpo}) and off-policy algorithms like DQN (\cite{mnih}) and DDPG \cite{ddpg}. In this technique, policies are represented as parameterized functions and to achieve exploration, the agent's policy is sampled from a set of policies perturbed by a noise distribution: $\widetilde{\theta}=\theta + \mathcal{N}(0,\sigma^2 I)$. The variance of the perturbation can be regulated to enhance exploration. In \cite{noisy}, the parameters of the noise distribution is learned using the gradients from the loss function of the RL algorithm.

\section{Perturbation-based exploration methods}
Our current work falls under the category of randomized exploration. In this section, we introduce the perturbed variants of the structured exploration techniques specified in section 2.1. We consider the perturbations both in policy and reward spaces and explore how they affect the policy performance in policy gradient algorithms.

\subsection{Exploration by sporadic intrinsic rewards}
The core idea behind sections 2.1.1 and 2.1.2 is that the intrinsic reward system employed by the agent could be different from that of the reward system of the environment (\cite{ryan}). The agent, based on its history in the domain,  might chose to prioritize exploring certain paths more than others to glean maximum information from the environment. It could potentially be advantageous for the agent to compromise its immediate returns for domain exploration, so it can leverage the information for maximizing its long-term returns. 

The learning of the agent's policy is deeply interposed with the reward function. When the reward function is perturbed with intrinsic reward bonuses, the agent's policy updates are now guided by the new reward function, although deviating from the intended objective, pushes the agent to explore different paths thus preventing its descent into a local minimum. These intrinsic reward bonuses which are introduced to perturb the reward function, are however obtained in a structured fashion in the works presented in 2.1, thus leading to a structured perturbation of the reward function.

In this work, we explore the effects of unstructured perturbations of the reward function. We sporadically introduce random intrinsic reward bonuses  into the environment to perturb the reward function. The transformed reward function is given by

  \[
  r_t =
  \begin{cases}
    r^{ext}     & \text{if $\nu < 0.5$}, \\
    r^{ext} + \beta \eta & \text{if $\nu > 0.5$}.
  \end{cases}
\]

where $\nu$, $\eta$ are random variables sampled from a Bernoulli distribution and $\beta$ is a variable that regulates the extent of perturbation and can be varied during the training process.

\subsection{Exploration by sporadic policy shaping}
Introducing intrinsic rewards to the environment can be considered as equivalent to reward shaping. This analysis can be similarly extended to the policy space where policies can be shaped in order to encourage exploration of different courses of actions across episodes of MDP. In fact entropy regularization is a form of policy shaping to prevent predictability of the agent's actions in a given state.

Consider the function approximator $\pi_m$ describing the policy of the agent. During execution, however, the agent deviates from the original policy by perturbing $\pi_m$ resulting in a behavior policy $\pi_b = \widetilde{\pi_m}$. In general, when the behavior policy deviates from the original policy, the learning is considered to be in off-policy mode. Swerving off-policy would require corrections to the policy gradient updates through incorporation of importance sampling factors to off-set the change in sampling (\cite{off-pac}). However when the perturbations are small, we can ignore the importance sampling factors as the behavior policy is in the neighbourhood of the target policy and is guided by the same parameters. The policy gradient updates of the behavior policy can be expressed in terms of parameters of target policy through reparametrization trick.

The function approximator for the policy is a deep neural network with a softmax layer at the end. In this study, we perturb the policy just before the softmax layer. Consider $f$ is the function approximator before the softmax layer, then the policy is represented as 
\begin{equation}
    \pi_m(\mathbf{s}) = f(\mathbf{s})
\end{equation}
Let the $\bm{\epsilon}$ be the array of perturbation factors. The behavior policy $\pi_b$ can be written as follows
\begin{equation}
    \pi'_m(\mathbf{s}) = \pi_m(\mathbf{s})  \cdot (1+\bm{\epsilon})
\end{equation}
\begin{equation}
    \pi'_m(\mathbf{s}, a) = \frac{\pi'_m(\mathbf{s}, a)}{\sum_{a} \pi'_m(\mathbf{s}, a)}
\end{equation}
\begin{equation}
    \pi_b(\mathbf{s}, a) = \textnormal{softmax}(\pi'_m(\mathbf{s}, a))
\end{equation}

We now describe how to derive the array of perturbation factors, $\bm{\epsilon}$, from prediction models for structured exploration. To enable the agent to explore efficiently we need to prioritize those actions in a state that the agent has chosen the least so far. To quantify the engagement of the agent with an action in a given state, we train autoencoder to encode a state-action pair each time it undertakes an action. Thus, the encoding error will be lowest for the action that the agent has taken many times and highest for the action that the agent has never chosen. The perturbation factors can now be defined as a ratio of encoding errors of all actions. 

Let $f_a$ be the autoencoder defined to encode state-action pairs. Let the $\mathbf{s}_a = (\mathbf{s},a)$ be the concatenation of state-action pair, $\mathbf{s}_a^e = f(\mathbf{s}_a)$ be the state-action encoding of $\mathbf{s}_a$.

\begin{equation}
    e_a = \Vert f(\mathbf{s}_a) - \mathbf{s}_a)\Vert^2
\end{equation}
\begin{equation}
   \epsilon_a = \frac{e_a}{\sum_a e_a}
\end{equation}

In case of sporadic policy shaping, we chose $\bm{\epsilon}$ to be an array of random values lying within a range of $(0,\eta)$.

\section{Experiments \& Results}
We conducted a series of experiments to test the efficacy of perturbation-based exploration techniques described in the previous section. We restricted our focus to policy-gradient class of algorithms: Proximal Policy Optimization (PPO), Actor Critic using Kronecker-Factored Trust Region (ACKTR), Advantage Actor Critic (A2C). 

In our experiments, we compared the two unstructured exploration techniques (sporadic intrinsic rewards, sporadic policy shaping) with the standard baselines to the above algorithms available in literature. For baselines, we considered Atari 2600 games (\cite{arcade}), Mujoco (\cite{mujoco}) and SparseCartpole (\cite{sp}), thus spanning complex input, discrete-action space, continuous control and sparse reward settings.  

\subsection{Comparison with PPO baselines}
We considered the PPO implementation from \cite{} as a framework for our evaluations. 
We compared our results with the standard PPO baselines from \cite{ppo}.
\subsubsection{Arcade learning environment}
To implement exploration with sporadic intrinsic rewards, we introduced random intrinsic rewards with lie in the range $[0,0.1]$ with a probability of $\eta = 0.5$. 
We implemented exploration with sporadic policy shaping as mentioned in section 3.2. We set $\epsilon=0.5$ in equation (5). The learning curves are provided as an average of 5 random seeds over 10 million timesteps. The hyperparameters for the experiments are listed in the appendix.

Figure (1) shows the comparison in performance of the two exploration techniques with the standard PPO baseline for six Atari 2600 games. In case of Frostbite, Alien, BeamRider remarkable performance benefit is achieved by both the exploration techniques. In Assault, the technique of sporadic intrinsic rewards has proved more effective while in the case of DemonAttack, sporadic policy shaping provided a performance benefit over the baseline. The performance curves for all the 47 games are provided in the appendix. The table contrasting the mean of the last hundred episodes for all the two exploration techniques and the baseline is also provided in the appendix. Out of 47 games, exploration by sporadic intrinsic rewards beat the baseline 23 times while the same feat is accomplished by sporadic policy shaping 19 times. In 18 of the 47 games, baseline remained the highest performing algorithm. 
\begin{figure*}[htbp]
\begin{subfigure}[b]{.33\linewidth}
\includegraphics[width=\linewidth]{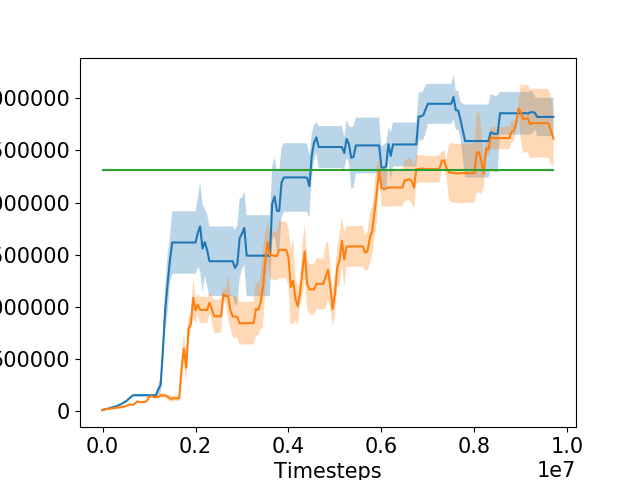}
\caption{Atlantis}\label{fig:mouse}
\end{subfigure}\hfill
\begin{subfigure}[b]{.33\linewidth}
\includegraphics[width=\linewidth]{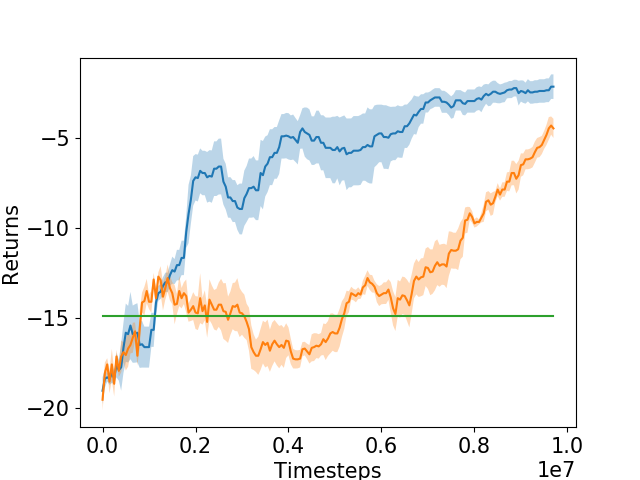}
\caption{DoubleDunk}\label{fig:gull}
\end{subfigure}\hfill
\begin{subfigure}[b]{.33\linewidth}
\includegraphics[width=\linewidth]{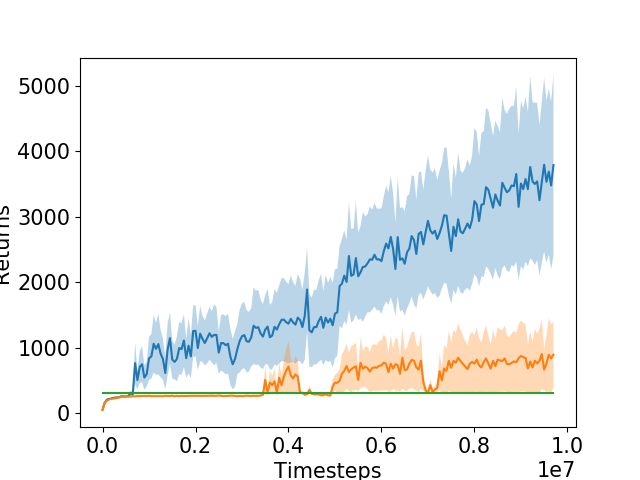}
\caption{Frostbite}\label{fig:tiger}
\end{subfigure}
\begin{subfigure}[b]{.33\linewidth}
\includegraphics[width=\linewidth]{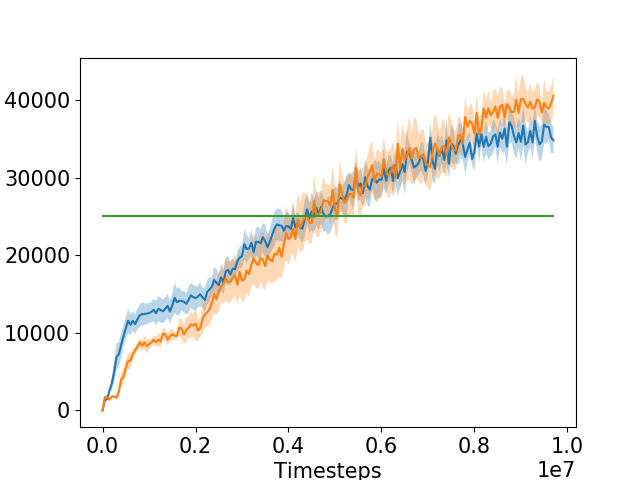}
\caption{RoadRunner}\label{fig:mouse}
\end{subfigure}\hfill
\begin{subfigure}[b]{.33\linewidth}
\includegraphics[width=\linewidth]{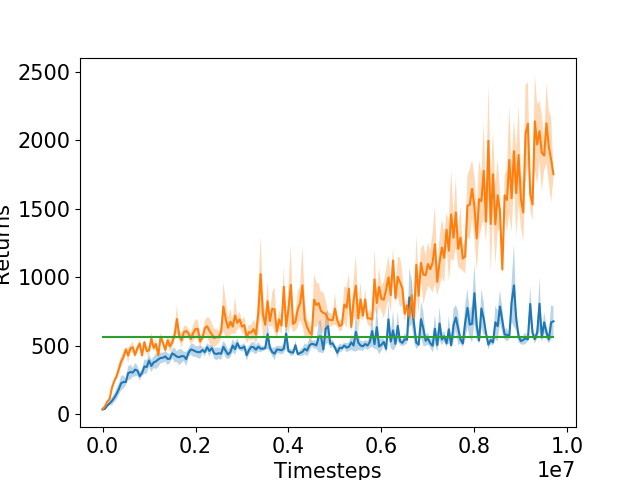}
\caption{Jamesbond}\label{fig:gull}
\end{subfigure}\hfill
\begin{subfigure}[b]{.33\linewidth}
\includegraphics[width=\linewidth]{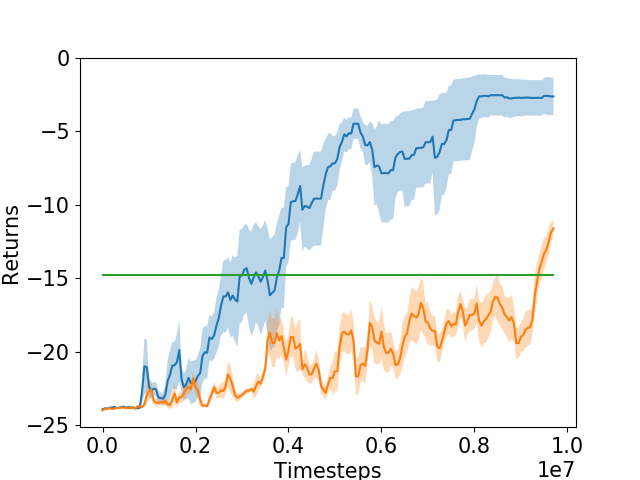}
\caption{Tennis}\label{fig:tiger}
\end{subfigure}
\caption{Comparison of sporadic intrinsic rewards (\protect\orangeline), sporadic policy shaping (\protect\blueline) and PPO baseline (\protect\greenline) on Atari 2600 games.}
\label{fig:animals}
\end{figure*}
\subsubsection{Continuous control environments}
We considered three Mujoco learning environments: HalfCheetah, Walker and Hopper. 
The intrinsic rewards were set to lie in the range $[0,0.01]$ with a probability of $\eta=0.5$. Average learning curves over 5 random seeds and over 1 million timesteps are provided in Figure 2. 

As shown, exploration by sporadic intrinsic rewards shows a benefit over the baseline in HalfCheetah and Hopper. In the case of HalfCheetah, two of the random seeds report over 100\% improvement over the baseline
\begin{figure*}[htbp]
\begin{subfigure}[b]{.33\linewidth}
\includegraphics[width=\linewidth]{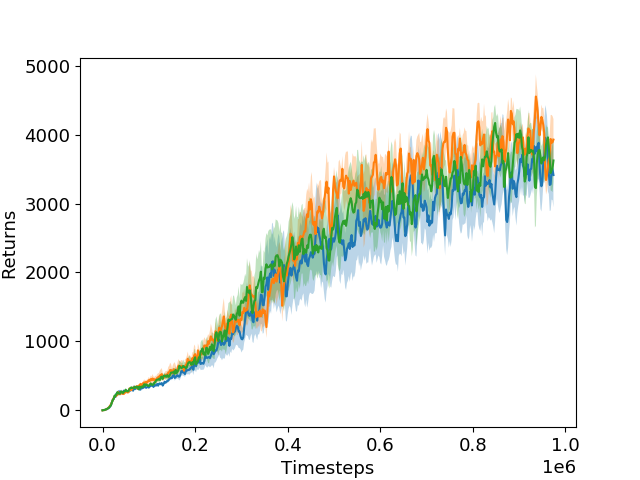}
\caption{Walker}\label{fig:mouse}
\end{subfigure}\hfill
\begin{subfigure}[b]{.33\linewidth}
\includegraphics[width=\linewidth]{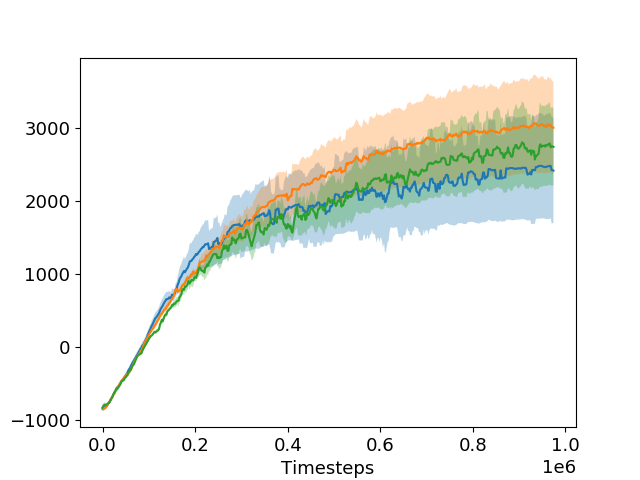}
\caption{HalfCheetah}\label{fig:gull}
\end{subfigure}\hfill
\begin{subfigure}[b]{.33\linewidth}
\includegraphics[width=\linewidth]{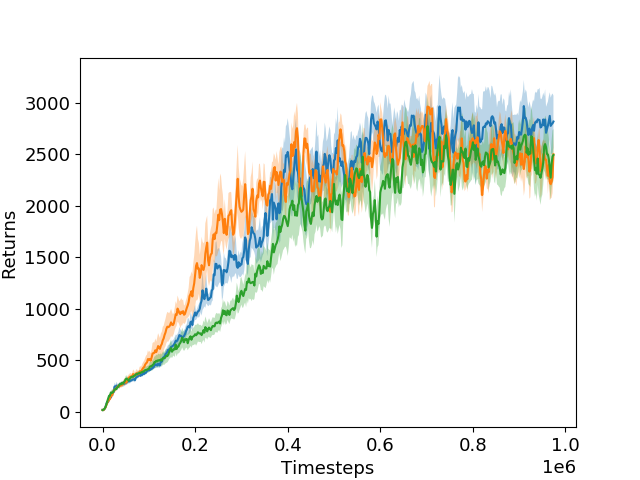}
\caption{Hopper}\label{fig:mouse}
\end{subfigure}
\caption{Comparison of sporadic intrinsic rewards (\protect\orangeline), sporadic policy shaping (\protect\blueline) and PPO baseline (\protect\greenline) on Mujoco continuous control environments.}
\end{figure*}

\subsection{Comparison with ACKTR baselines}
Exploration with sporadic intrinsic rewards were implemented as described in the case of PPO baselines. The baselines that are derived from \cite{acktr} are provided for a single seed. In this section, the learning curves in this experiment are also provided as a mean out of 3 random seeds. The performance curves for six Atari 2600 games are shown in Figure 3. Exploration with policy shaping is skipped on constraints on time and will be included in the final version. As shown in the figure, Asterix and Jamesbond show remarkable performance improvement upon introduction of sporadic intrinsic rewards. 

The results (tables, curves) for the set of 33 games are provided in the appendix. Out of 33 games, introducing unstructured intrinsic rewards provides a performance boost in 16 games.

\begin{figure*}[htbp]
\begin{subfigure}[b]{.33\linewidth}
\includegraphics[width=\linewidth]{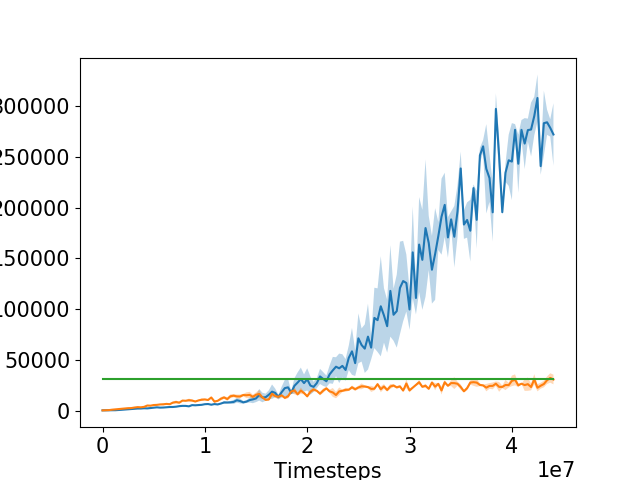}
\caption{Asterix}\label{fig:mouse}
\end{subfigure}\hfill
\begin{subfigure}[b]{.33\linewidth}
\includegraphics[width=\linewidth]{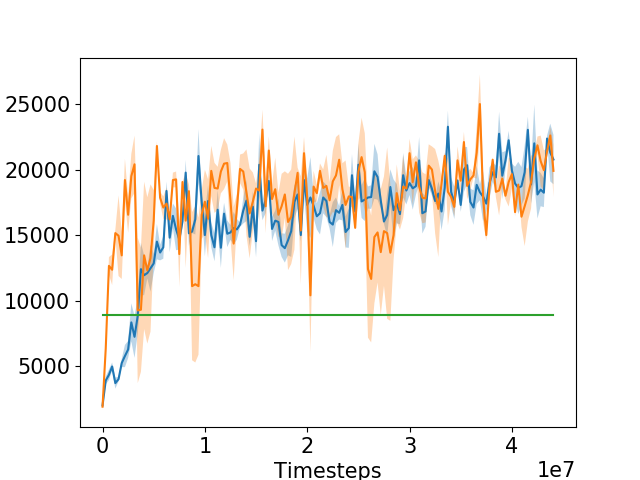}
\caption{BattleZone}\label{fig:gull}
\end{subfigure}\hfill
\begin{subfigure}[b]{.33\linewidth}
\includegraphics[width=\linewidth]{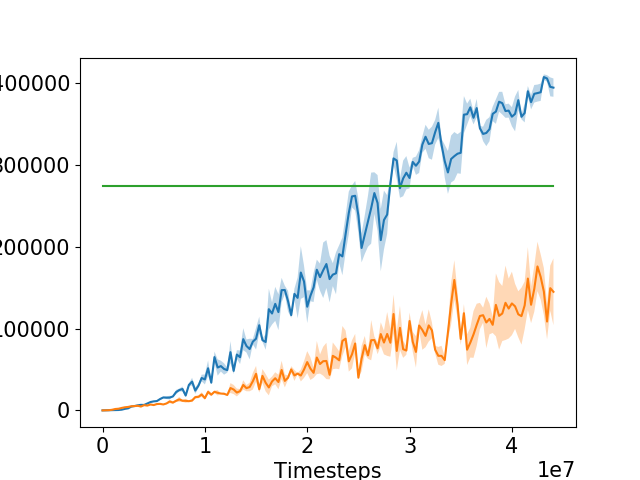}
\caption{DemonAttack}\label{fig:tiger}
\end{subfigure}
\begin{subfigure}[b]{.33\linewidth}
\includegraphics[width=\linewidth]{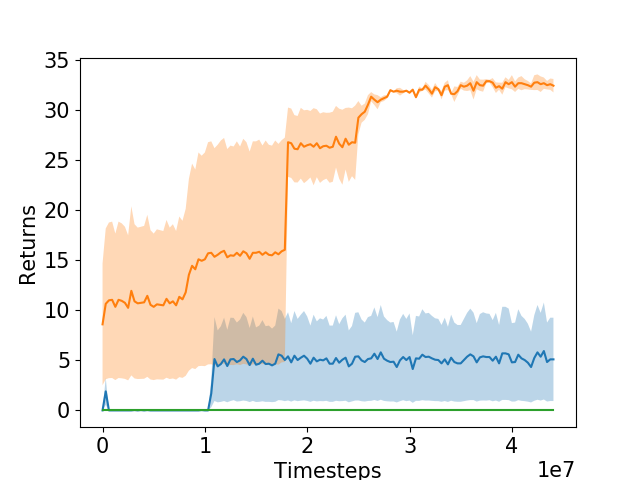}
\caption{Freeway}\label{fig:mouse}
\end{subfigure}\hfill
\begin{subfigure}[b]{.33\linewidth}
\includegraphics[width=\linewidth]{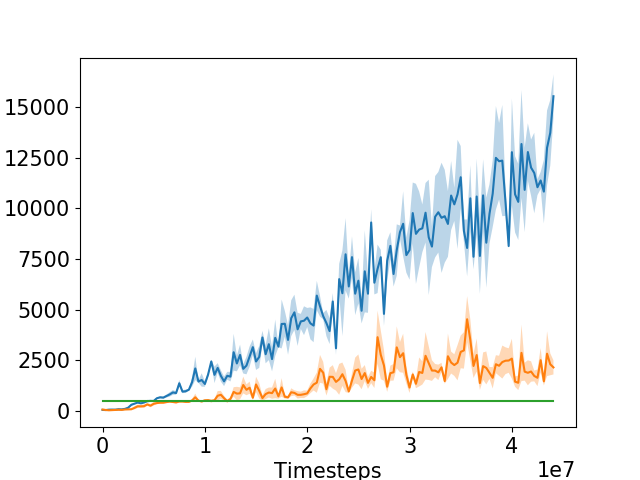}
\caption{Jamesbond}\label{fig:gull}
\end{subfigure}\hfill
\begin{subfigure}[b]{.33\linewidth}
\includegraphics[width=\linewidth]{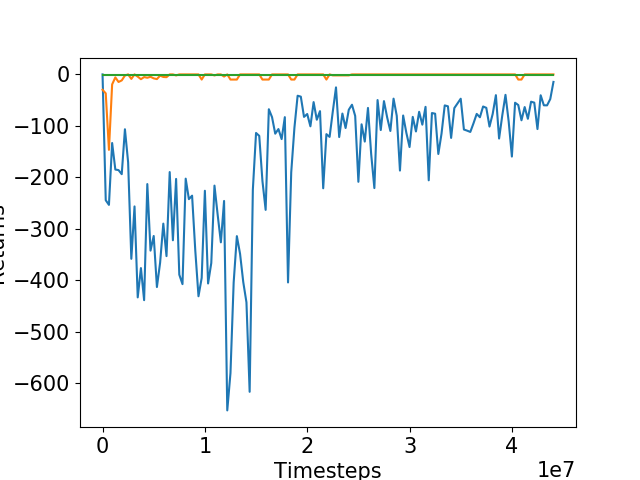}
\caption{Pitfall}\label{fig:tiger}
\end{subfigure}
\caption{Comparison of sporadic intrinsic rewards (\protect\orangeline), sporadic policy shaping (\protect\blueline) and ACKTR baseline (\protect\greenline) on Atari 2600 games. }
\label{fig:animals}
\end{figure*}

\subsection{Comparison with A2C baselines}
\begin{figure}[h]
  \begin{center}
    \includegraphics[width=0.4\textwidth]{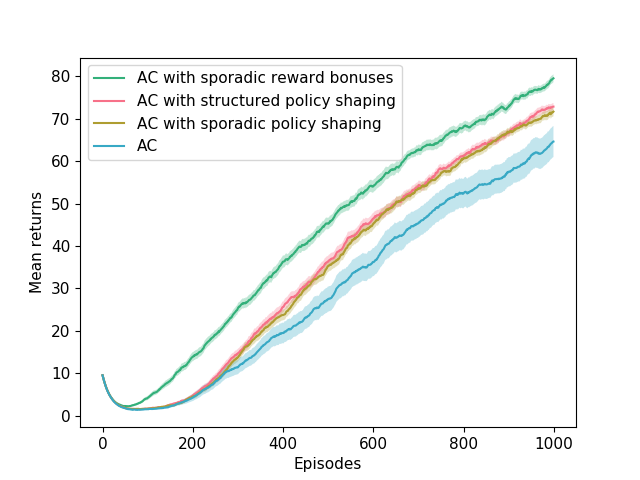}
  \end{center}
  \caption{Effect of unstructured and structured exploration on Advantage Actor Critic framework demonstrated on SparseCartpole task.}
\end{figure}

In this experiment, we consider the effects of both structured and unstructured exploration on the actor-critic learning framework. To make the task of exploration more challenging, we consider a sparse reward task, SparseCartpole (\cite{sp}), where the reward is provided to the agent only if the pole reaches a target height. 

In addition to the two unstructured exploration techniques, we also considered the structured policy shaping introduced in section 3.2. A three-layered neural network with 128 hidden neurons is used to parameterize the policy. Figure 4 shows the four learning curves including the baseline actor-critic version. Hyperparameters are tuned separately for each algorithm using a grid-search. As the figure demonstrates, all the four exploration techniques provide a significant improvement from the baseline both in terms of mean expected return and variance of the expected returns across episodes. Interestingly, in this scenario, structured policy shaping provides only a statistically insignificant improvement over its unstructured variant.

\section{Limitations}
In this work, we show considerable evidence that unstructured exploration techniques provide significant improvement over the baseline algorithms. However, the comparison of unstructured exploration, specifically the effect of sporadic intrinsic rewards, with that of other bonus-based exploration techniques in the literature (\cite{bonus}, \cite{count-based}, \cite{unifying}) is not provided, which is a major limitation of the work. Providing a thorough comparison of these techniques would enable us to disassociate the effect of perturbed reward space in the structured bonus-based exploration techniques. 

On account of low perturbation in the policy space and the reward space, we tuned the hyperparameters in the neighbourhood of the hyperparameters of the baseline. If we assume the baselines presented in \cite{ppo}, \cite{acktr} are rigorous, then any additional hyperparameter tuning on our side, would only serve to improve the results of our experiments. In our work, we have demonstrated that there is a significant improvement in performance upon using unstructured exploration techniques, however, the extent of improvement might vary upon further hyperparameter tuning. 

\section{Conclusion}
In this work, we introduced two perturbation-based exploration techniques for deep reinforcement learning. In the first of these techniques, by introducing random intrinsic rewards, we derived an unstructured variant of bonus-based exploration methods. This technique, with nominal overhead in computation, is shown to provide remarkable improvement in performance over the baseline in several games of the arcade learning environment. 

We also introduced policy shaping as a potential structured exploration technique and further derived its unstructured variant. We provided the comparision of the two exploration techniques on PPO, ACKTR, A2C frameworks applied to the domains of Atari2600, Mujoco and SparseCartpole, and demonstrated significant improvement over the baseline in all of the above settings. Our results emphasizes the relevance of perturbation-based exploration methods and the need to justify the structured exploration techniques by providing a comparison against their perturbed variants.

\begin{table*}[ht]
\centering

\begin{tabular}[t]{lccc}
\hline
&PPO & PPO with SR & PPO with PS\\
\hline
Alien&\textbf{1850.3}&1700.6&1827.2\\
Amidar&\textbf{674.6}&576.3&568.6\\
Assault&4971.9&\textbf{5297.5}&3715.9\\
Asterix&4532.5&4171.2&\textbf{4968.3}\\
Asteroids&2097.5&\textbf{2349.4}&1644.4\\
Atlantis&2311815&\textbf{2677784.4}&\textbf{3001185.0}\\
BankHeist&\textbf{1280.6}&1203.2&949.14\\
BattleZone&17366.7&\textbf{23871.7}&14663.5\\
BeamRider&1590.0&\textbf{3812.2}&\textbf{2093.1}\\
Bowling&\textbf{40.1}&32.7&15.6\\
Boxing&\textbf{94.6}&91.3&93.5\\
Breakout&274.8&260.0&\textbf{387.2}\\
Centipede&\textbf{4386.4}&4074.1&3769.9\\
ChopperCommand& \textbf{3516.3}&707.0&680.5\\
CrazyClimber&110202.0 &109066.7&\textbf{120332.1}\\
DemonAttack&11378.4&\textbf{11545.3}&8066.7\\
DoubleDunk&-14.9&\textbf{-3.9}&\textbf{-2.4}\\
Enduro&758.3&747.4&\textbf{1160.1}\\
FishingDerby&17.8&\textbf{24.7}&13.1\\
Freeway&\textbf{32.5}&32.0&30.0\\
Frostbite&314.2&\textbf{1064.7}&\textbf{2689.0}\\
Gopher&2932.9&\textbf{3079.4}&2516.0\\
Gravitar&\textbf{737.2}&311.7&578.8\\
IceHockey&-4.2&-4.54&\textbf{-4.1}\\
Jamesbond&560.7&\textbf{1854.8}&\textbf{623.2}\\
Kangaroo&\textbf{9928.7}&207.3&4559.7\\
Krull&7942.3&\textbf{8662.0}&\textbf{9270.6}\\
KungFuMaster&23310.3&\textbf{24523.6}&\textbf{30304.6}\\
MsPacman&2096.5&\textbf{2192.0}&\textbf{2106.0}\\
NameThisGame&\textbf{6245.9}&5530.7&5450.0\\
Pitfall&\textbf{-32.9}&-214.7&-82.8\\
Pong&\textbf{20.7}&20.2&18.37\\
PrivateEye&\textbf{69.5}&61.0&25.0\\
Qbert&\textbf{14293.3}&13626.5&6546.3\\
Riverraid&8393.6&\textbf{8803.2}&6940.0\\
RoadRunner & 25076.0&\textbf{38865.3}&\textbf{36231.2}\\
Robotank&5.5&\textbf{10.8}&\textbf{9.0}\\
Seaquest&1204.5&\textbf{1387.5}&\textbf{1254.1}\\
SpaceInvaders&\textbf{942.5}&846.8&875.1\\
StarGunner&32689.0&\textbf{36088.3}&18479.8\\
Tennis&-14.8&\textbf{-12.1}&\textbf{-1.2}\\
TimePilot&4342.0&\textbf{5219.7}&\textbf{7070.0}\\
Tutankham&\textbf{254.4}&112.4&143.0\\
UpNDown&95445.0&\textbf{311425.8}&\textbf{313143.1}\\
VideoPinball&\textbf{37389.0}&29207.1&30301.5\\
WizardOfWor&4185.3&\textbf{7992.2}&3735.6\\
Zaxxon&5008.7&4023.0&\textbf{5336.0}\\
\hline
\end{tabular}
\vspace{10pt}
\caption{Table of comparisons between PPO baseline, PPO with Sporadic Rewards (SR), PPO with Policy Shaping (PS) on Atari 2600 games}
\end{table*}%

\begin{table*}[ht]
\centering
\begin{tabular}[t]{cc}
\hline
Hyperparameter & Value \\
\hline
Horizon (T) & 128 \\
Adam stepsize & $2.5 \times 10^{-4} \times \alpha$\\
Number of PPO epochs & 4\\
Minibatch size & 4\\
Discount parameter ($\gamma$) & 0.99\\
GAE parameter ($\lambda$) & 0.95\\
Number of actors & 8 \\
Clipping parameter ($\epsilon$) & $0.1 \times \alpha$ \\
Value loss coefficient & 0.5 \\
Entropy coefficient & 0.01 \\

\end{tabular}
\vspace{10pt}
\caption{Hyperparameters for arcade learning environment as applied to \cite{pytorch}}
\end{table*}

\begin{table*}[ht]
\centering
\begin{tabular}[t]{lcc}
\hline
&ACKTR & ACKTR with SR \\
\hline
Alien & \textbf{3197.1} & 3072.7\\
Amidar & \textbf{1059.4} & 839.8  \\
Assault & 10777.7 & \textbf{15932.4}\\
Asterix & 31583.0 & \textbf{297435.0}\\
Asteroids & \textbf{34171.0} & 3886.0\\
Atlantis & \textbf{3433182.0} & 3085143.3\\
BankHeist & \textbf{1289.7} & 1240.3 \\
BattleZone & 8910.0 & \textbf{18500.0} \\
BeamRider & 13581.4 & \textbf{15089.7} \\
Bowling & 24.3 & \textbf{26.6} \\
Boxing & \textbf{1.45} & 0.9 \\
Breakout & \textbf{735.7} & 558.3\\
Centipede & 7125.8 & \textbf{7679.6} \\
CrazyClimber & \textbf{150444.0} & 135876.7 \\
DemonAttack & 274176.7 & \textbf{409639.2} \\
DoubleDunk & \textbf{-0.54} & -2.4\\
FishingDerby & 33.7 & \textbf{34.5}\\
Freeway & 0.0 & \textbf{5.1}\\
Gopher & 47730.0 & \textbf{48156.0}\\
IceHockey & -4.2 & \textbf{-3.6} \\
Jamesbond & 490.0 & \textbf{12926.7}\\
Kangaroo & \textbf{3150.0} & 2495.0 \\
Krull & \textbf{9686.9} & 9596.7\\
KungFuMaster & 34954.0 & \textbf{37180.0}\\
Pitfall & \textbf{-1.1} & -90.9\\
Pong & \textbf{20.9} & 18.6\\
Qbert & 23151.5 & \textbf{24327.5}\\
Riverraid & 17762.8 & \textbf{17769.7}\\
RoadRunner & \textbf{53446.0} & 41240.0\\
Robotank & 16.5 & \textbf{22.6}\\
Seaquest & \textbf{1776.0} & 1708.0\\
Solaris & \textbf{2368.6} & 1498.0 \\
SpaceInvaders & 19723.0 & 2747.5\\
\hline
\end{tabular}
\vspace{10pt}
\caption{Table of comparisons between ACKTR and ACKTR with Sporadic Rewards (SR) on Atari 2600 games.}
\end{table*}%
\bibliography{aaai}

\end{document}